 \documentclass[final,5p,times,twocolumn,authoryear]{elsarticle}

\usepackage{amssymb}
\usepackage{lipsum}
\usepackage{booktabs}
\usepackage{forest}
\usepackage{lipsum}  
\usepackage{hyperref}
\usepackage{cleveref}
\usepackage{url}

\crefformat{section}{\S#2#1#3} % see manual of cleveref, section 8.2.1
\crefformat{subsection}{\S#2#1#3}
\crefformat{subsubsection}{\S#2#1#3}

\setcitestyle{parenthesis,citesep={,}}
\journal{Journal of Web Semantics}

\begin{document}

\begin{frontmatter}



%\title{Knowledge Graphs for LLMs: An NLP Perspective}
%\title{Knowledge Graphs and Hallucinations in Large Language Models: An NLP Perspective}
\title{Knowledge Graphs, Large Language Models, and Hallucinations: An NLP Perspective}


\author[first]{Ernests Lavrinovics\corref{cor1}}
\cortext[cor1]{Email: elav@cs.aau.dk}

\author[first]{Russa Biswas}
\author[first]{Johannes Bjerva}
\author[second]{Katja Hose}
\affiliation[first]{organization={Aalborg University, Department of Computer Science}, 
            city={Copenhagen},
            country={Denmark}}
\affiliation[second]{organization={TU Wien, Institute of Logic and Computation},
            city={Vienna},
            country={Austria}}

\begin{abstract}
Large Language Models (LLMs) have revolutionized Natural Language Processing (NLP) based applications including automated text generation, question answering, chatbots, and others. However, they face a significant challenge: hallucinations, where models produce plausible-sounding but factually incorrect responses. This undermines trust and limits the applicability of LLMs in different domains. Knowledge Graphs (KGs), on the other hand, provide a structured collection of interconnected facts represented as entities (nodes) and their relationships (edges). In recent research, KGs have been leveraged to provide context that can fill gaps in an LLM’s understanding of certain topics offering a promising approach to mitigate hallucinations in LLMs, enhancing their reliability and accuracy while benefiting from their wide applicability. Nonetheless, it is still a very active area of research with various unresolved open problems. In this paper, we discuss these open challenges covering state-of-the-art datasets and benchmarks as well as methods for knowledge integration and evaluating hallucinations. In our discussion, we consider the current use of KGs in LLM systems and identify future directions within each of these challenges. % with a goal to pave the way for future research.
\end{abstract}



\end{frontmatter}

\section{Introduction}
\label{introduction}
Large Language Models (LLMs) are anticipated to have a substantial impact on domains such as law, cyber security, education, and healthcare due to their ability to generalize well on language technology tasks, such as text summarization, question-answering (QA), and others \citep{augenstein2023factuality}. A major flaw that prevents widespread deployment of LLMs is factual inconsistencies, also referred to as hallucinations, which impair trust in AI systems and even pose societal risks in the form of generating convincing false information \citep{augenstein2023factuality, puccetti2024ai}. 

\begin{figure}[!htb]
    \centering
    \includegraphics[width=0.85\linewidth]{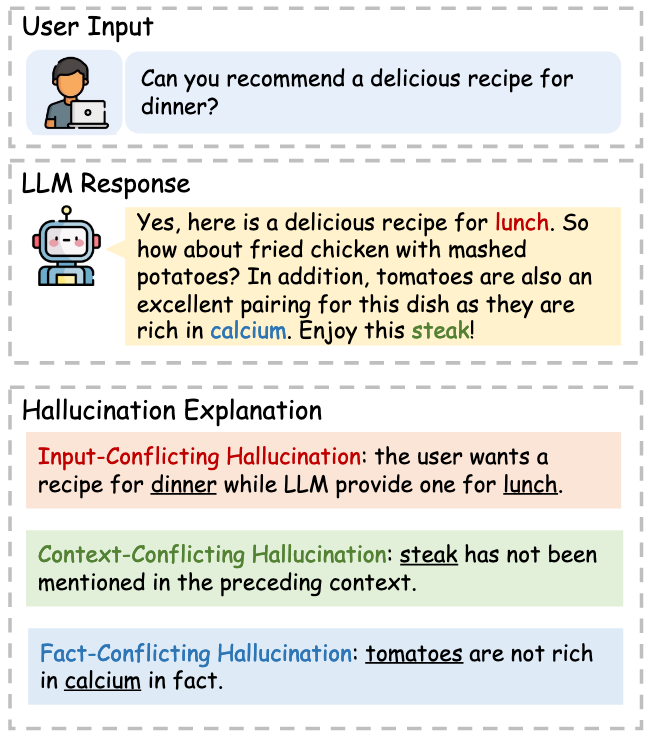}
    \caption{Example of different types of hallucinations occurring in the same output \citep{zhang2023siren}.}
    \label{fig:halluc_eg}
\end{figure}

Hallucinations are a multifaceted problem as there are conceptually different types such as hallucinations with respect to world knowledge, self-contradictions, with respect to prompt instructions or given context \citep{huang2023survey, zhang2023siren}, see Figure \ref{fig:halluc_eg}. While \cite{10569238} points out that hallucinations can be useful for brainstorming or generating artwork, they are a limiting factor for contexts where factuality is a priority, including use cases that require large-scale text processing, such as question answering, information retrieval, summarization, and recommendations. % or any other contexts where factuality is prioritized then hallucinations are indeed a limiting factor. 
Therefore, research towards robust methods of generating consistent output with LLMs given factual and informative inputs is still an active and ongoing direction. A naïve approach to updating LLM internal knowledge is through means of retraining the model which is a time-consuming and expensive process. 

Recent research \citep{pan2024unifying, russaKgSurvey} has identified Knowledge Graphs (KGs) as relevant structured information of knowledge for factual grounding that LLMs can be synergized with and conditioned on to improve general factual consistency of an LLM's output. KGs are structured representations of knowledge in a graph-like structure consisting of entities, relationships, and attributes that encode factual information about real-world objects in a machine-readable format. KGs can alleviate the need for full retraining by providing a factual basis that can be utilized during inference or post-generation. This is especially critical in use cases where knowledge evolves fast and time and computational resources are limited.

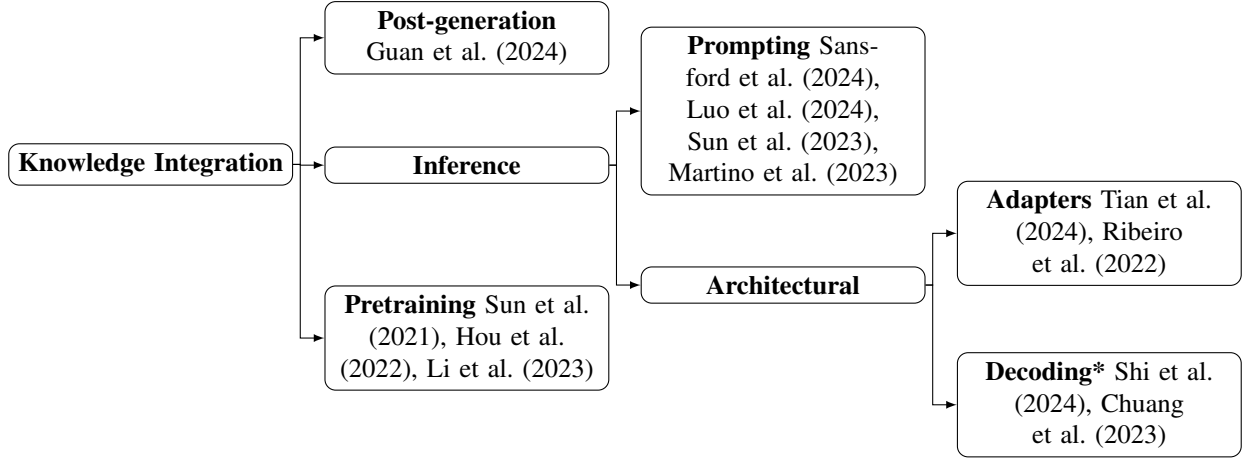
\begin{figure*}[!htb]
\centering
\begin{forest}
  for tree={
    grow=east,
    draw,
    edge={-latex},
    rounded corners,
    node options={align=center, text width=3.5cm}, % Limits the width of each node
    s sep=25pt, % Increases horizontal spacing between nodes
    l=1cm, % Adds vertical spacing
    edge path={\noexpand\path [draw, \forestoption{edge}]
      (!u.parent anchor) -- +(56pt,0) |- (.child anchor)\forestoption{edge label};}
  }
    [\textbf{Knowledge Integration}
            [\textbf{Pretraining} \cite{sun2021ernie, hou2022adapters, li2023knowledge}]
            [\textbf{Inference} 
                [\textbf{Architectural}
                    [\textbf{Decoding*} \cite{shi2023trusting, chuang2023dola}]
                    [\textbf{Adapters} \cite{tian-etal-2024-kg, ribeiro-etal-2022-factgraph}]
                ]
                [\textbf{Prompting} \cite{sansford2024grapheval, luo2023reasoning, sun2023think, martino2023knowledge}]
            ]
            [\textbf{Post-generation} \cite{guan2024mitigating}]
        ]
\end{forest}
\caption{Our categorization of different stages at which external knowledge can be integrated in an LLM to mitigate hallucinations. *Decoding does not explicitly use KGs although it can be used to prioritize in-context knowledge (such as KG metadata).}
\label{fig:hallucination_mitigation}
\end{figure*}

Previous work explores methods for integrating and conditioning LLMs on factual inputs. A related survey paper~\citep{zhang2023siren} presents a comprehensive overview of different types of hallucinations in LLMs and mitigation models, whereas in this paper we focus on the KG-based hallucination mitigation models. We propose a categorization of different knowledge integration models based on their underlying architecture. Figure \ref{fig:hallucination_mitigation} depicts the categorization of different stages at which additional information can be included to boost factuality.

The performance of these solutions varies depending on the methodological knowledge injection as well as the underlying LLM used. Furthermore, the evaluation of hallucinations is a complex problem in itself since for generative tasks it is necessary to evaluate the semantics of the output. Keeping this in mind, there are metrics, such as BERTScore~\cite{zhang2019bertscore}, and BARTScore~\cite{yuan2021bartscore} that evaluate semantic similarity between two pieces of text, e.g., LLM output and reference text. Additionally, textual entailment models can be used to classify whether a part of a hypothesis (LLM output) entails or contradicts a given premise (factual knowledge). 

Methods such as BERTScore, BARTScore, and entailment models process a whole hypothesis holistically. 
However, hallucinations can be subtle, and even a single incorrect word can result in a large semantic mismatch.
Hence, these methods tend to fail to accurately describe hallucinations, as they are not able to capture granular word-level details.

Therefore, in order to discover reliable methods for hallucination mitigation there is a need for robust evaluation, which is currently not present although being an active research direction. There are numerous benchmarks proposed for evaluating hallucination detection models as shown in Table \ref{tab:datasets}. However, the majority of them focuses on response-level granularity. Considering the subtleness of hallucinations, it is necessary to have a finer granularity at , for example, sentence-level as proposed in the FELM benchmark \cite{zhao2024felm} or even span-level as per MuShroom-2025 \cite{shroom2025}.
The MuShroom-2025 shared task aims to enable research by proposing an open-sourced dataset for span-level hallucination detection, meaning that the task requires participants to identify exact portions of the text in which hallucinations occur. If the problem of hallucination detection can be solved at scale, then this provides a stable foundation for discovering effective and scalable methods for mitigating certain types of hallucinations.

In summary, this position paper proposes a categorization of knowledge integration methods that use KGs as per Figure \ref{fig:hallucination_mitigation} and consolidates available resources in Table \ref{tab:datasets}. Furthermore, we argue for the importance of the following open research directions in which KGs can play a critical role:

\begin{enumerate}
    \item Robust detection of hallucinations with a fine-grained overview of particular hallucinatory text spans
    \item Effective methods for integrating knowledge in LLMs that move away from textual prompting
    \item Evaluation of factuality in a multiprompt, multilingual, and multitask space for an in-depth analysis of model performance
\end{enumerate}
The remainder of this paper is structured as follows:  
Section~\ref{sect:res} discusses modern datasets and benchmarks, %, and outlines expected requirements for more robust resources;
Section~\ref{sect:feasibility} discusses %and outlines 
the feasibility of mitigating hallucinations, 
Section~\ref{sect:hallcDet} gives an overview of hallucination detection methods, %and discusses inherent limitations and research gaps,  
Section~\ref{sect:knowinject} discusses how additional knowledge can be integrated to mitigate hallucinations, and 
%outlines three main stages at which additional knowledge can be integrated for hallucination mitigation, as well as emphasizes problem of multilinguality, limitations and tradeoffs when integrating external knowledge, 
Section~\ref{sect:hallcEval} outlines current methods for evaluating hallucinations. %, as well as research gaps and limitations of methods.
Finally, Section~\ref{sect:conclusions} summarizes identified research gaps 

\section{Available Resources for Evaluating Hallucinations}
\label{sect:res}
Considering the boom of LLMs in recent years, evaluation of hallucinations has become increasingly important due to the anticipated high value that LLMs can provide for problem solving. This has sparked an increase in dedicated evaluation datasets and benchmarks, Table \ref{tab:datasets} shows an overview. 

For the LLM hallucination evaluation to be holistic, we argue that evaluation needs to broadly cover different domains as well as different tasks to test for different types of hallucinations. One of the major objectives for LLMs to be useful for practical applications is generalizability to multiple domains. Table \ref{tab:datasets} reveails that many of the datasets cover evaluation on a multi-domain basis such as law, politics, medical, science and technology, art, finance, and others. 

Most of the datasets are primarily focused towards evaluating hallucination detectors that output information about whether a hallucination is present in a piece of text on a response, sentence, or span level. While this does not explicitly model hallucination evaluation for a given LLM, the data points can be re-purposed for hallucination evaluation.

It is also evident from Table \ref{tab:datasets} that most of the datasets are actually benchmarks, therefore not providing dedicated training splits that can be used to train parametric knowledge integration models. All datasets in Table \ref{tab:datasets} except SemEval2025-MuShroom are available only in English therefore neglecting any kind of multilingual evaluation, and thus limiting the accessibility of LLM technology. Additionally, knowledge sub-graphs as additional context are not a popular feature of any of the datasets, therefore again limiting the methods that the evaluation and training can be performed on. % given models and methods. 
Given either textual, context, or Web pages as a resource, the primary use case is the evaluation of models based on retrieval augmented generation (RAG) using unstructured text.

Furthermore, previous work \cite{whatSota} outlines the need for a multiprompt evaluation, as the output of LLMs can depend on the phrasing of the input.The only dataset that evaluates such robustness and consistency is \cite{rahman2024defan} by accompanying each question-answer datapoint with 15 different paraphrasings of the same question.

Therefore, we conclude that there are many gaps in high-quality evaluation and training resources that have to be closed before they can be used for hallucination evaluation and mitigation, especially through using KGs.

\begin{table*}[!htb]
  \centering
  \resizebox{\textwidth}{!}{%
  \begin{tabular}{p{2,5cm}|lp{4,5cm}lllllll}
\toprule
Name    & Domain  & Task types & Splits                     & Sub-tasks (n) & Size     & Ext.Knowledge & Evaluation  &  Granularity \\
\midrule
MedHalt \cite{umapathi2023med} & Medical & Hallc. Evaluation (Reasoning, IR) & Train/Val/Test & 7 & 19k & None      & Accuracy, PWS &  Response \\
HaluEval \cite{li2023halueval} & General & Hallc. Detection (QA, Summ., Dialogue) & Test* & 4 & 35k & Context & Accuracy  & Response \\
Shroom SemEval 2024 \cite{mickus-etal-2024-semeval} & General    & Hallc. Detection (MT, PG, DM) & Train/Val/Test & 3 & 12k & None    &  Accuracy, calibration&  Response \\
MuShroom SemEval 2025 \cite{shroom2025} & General    & Hallc. Detection (QA) & Train/Val/Test & 1 & N/A & None   &  Accuracy, calibration&  Span \\
TruthfulQA \cite{lin2021truthfulqa} & Multi-domain   & Hallc. Evaluation (QA) & Test   & 3 &  817 & Web  & GPT-Judge & Response \\
FELM \cite{zhao2024felm} & Multi-domain   & Hallc. Detection (Reasoning, QA, recommendations) & Test &  5  & 169 & Web & F1, accuracy & Segment \\
&         &                                 &                       &                &  \\
HaluBench \cite{ravi2024lynxopensourcehallucination} & Multi-domain & Hallc. Detection (QA) & Test &  1  & 15k  & Context & Accuracy & Response \\
&         &                                 &                       &                &  \\
DefAn \cite{rahman2024defan} & Multi-domain & Hallc. Evaluation (QA) & Test &  1  & 75k & None & FCH, PMH, RC &  Response \\
SimpleQA \cite{openAiSimpleQA} & Multi-domain & Hallc. Evaluation (QA) & Test &1  & 4.3k & Web & F1 &  Response \\
\bottomrule
\end{tabular}
  }
  \caption{Overview of available resources for hallucination detection and evaluation. Task abbreviations: Machine Translation (MT), Paraphrase Generation (PG), Definition Modeling (DM), Summarization (summ.), Question-Answering (QA), Information Retrieval (IR). \textbf{*}HaluEval test split is based on the \textit{train} split from datasets such as HotpotQA, CNN/DailyMail and OpenDialogueKG. All datasets are in English except MuShroom SemEval 2025 (language n=10).}
  \label{tab:datasets}
\end{table*}

\section{Feasibility of Hallucination Mitigation}
\label{sect:feasibility}
Previous works criticize LLMs based on the hallucination phenomena and outline through defined formalisms that LLMs will not be 100\% free from the risk of hallucinations \cite{xu2024hallucination, banerjee2024llms}. On the other hand, \cite{xu2024hallucination} outlines that access to external knowledge can be an effective mitigator of hallucinations although the scalability remains unclear. This raises essentially two requirements for improving reliability of LLM systems, namely: (1) enabling output interpretability, allowing the end-user to scrutinize the output due to proneness of hallucinations; (2) conditioning an LLM on a reliable external knowledge source for mitigating hallucinations.

To this end, KGs are useful under the assumption that the knowledge graph triples are factually correct with respect to the user query. If an LLM uses the KG triples effectively, then its output can be mapped back to the knowledge graph that information originates from so it can be cross-checked and scrutinized as needed.
% TODO: review \cite{agrawal-etal-2024-knowledge}

\section{Detection of Hallucinations}
\label{sect:hallcDet}
Hallucination detection is the task of determining whether a particular piece of text generated by an LLM contains any form of hallucinations. This is a difficult task due to the multi-faceted nature of the problem.

GraphEval \cite{sansford2024grapheval} proposes a two-stage method for detecting and mitigating hallucinations with respect to a given textual context as ground-truth. The detection methodology proposes extracting atomic claims from the LLM output as a sub-graph by LLM-prompting and comparing each triple's entailment to the given textual context. 

Similarly, \cite{factalign} extracts KG subgraphs between source and generated text based on named entities (organizations, places, people, etc.) to then compare the alignment between the two graphs. Classification of the hallucination is done by thresholding the alignment. If a KG is built around named entities, this could lead to information loss on more abstract concepts, therefore improvements can be made towards more comprehensive relation extraction. KGR \cite{guan2024mitigating} also performs hallucination detection through designated system modules for claim extraction, fact selection, and verification. Fact selection relies on the information extraction abilities of LLM's, which in themselves are prone to hallucinations, therefore this raises a problem of effective and reliable query generation based on the given claims. Fleek \cite{fatahi-bayat-etal-2023-fleek} is a system demonstration aimed for fact-checking. The authors extract relevant claims as structured triples and verify them against a KG or a Web search engine by generating questions with a separate LLM based on the extracted claims.

The general trend of evaluating claims on an atomic level by representing them as KG structures enables output interpretability by allowing to return the inconsistent triples. This enables highlighting of problematic text spans and scrutiny of the output. Manual evaluation can also benefit understanding problematic use cases. However, none of these methods demonstrate results that would suggest the task at hand being solved, and the limited evaluation datasets also do not provide an insight into how truly generalizable these methods are. 

Considering the inherent limitations of LLMs, we raise skepticism over the scalability and robustness of methods that use multi-stage pipelines for extracting and validating claims if fully relying on LLM prompting to make the judgements at each processing step. We therefore call for the need for further evaluation of such methods of more diverse datasets or at least reporting on fine-grained analysis of each submodule's error rates and performance. We also advocate for lines of research that perform the task without primary reliance on LLMs.

Furthermore, it is unclear how the mixture of hallucination detection methods scales when combined with other fundamentally different approaches, e.g., LLM uncertainty for hallucination detection \cite{zhang2023enhancing}. We therefore propose to investigate the synergy between uncertainty and KG-based hallucination detection.

\section{Methods for Integrating Knowledge from KGs in LLMs}
\label{sect:knowinject}

Many previous methods explore integrating external knowledge as part of a larger LLM system as depicted in the categorization in Figure \ref{fig:hallucination_mitigation}. Knowledge from KGs can be integrated at different stages of an LLM system, whether its pretraining, inference, or post-generation. 
In the following, we discuss the methodological qualities of each of these stages. 
%The following sections scope-in on each of these stages and discuss the methodological qualities.
% TODO: Review \cite{rho} Rho paper.

%\subsection{Knowledge in Pretraining}
\bigskip
\textbf{Knowledge in Pretraining. }
Factually informed pretraining has been explored through incorporating KG triples as part of the training pipeline \cite{sun2021ernie}. The contribution proposes a methodology for fusing KG triples with raw text input by a masked entity prediction task. For a sentiment analysis task \cite{li2023knowledge} information from KGs is combined with text through a dedicated fusion module. Additionally, adapter-based techniques \cite{hou2022adapters} have been proposed that encode knowledge from KGs acting as low-parameter add-ons to an LLM architecture. This creates factually aware neural modules that, when plugged into a larger LLM architecture, suggest to boost factuality.

%\subsection{Knowledge During Inference}
\bigskip
\textbf{Knowledge During Inference. }
A common naïve method to integrate external knowledge is through prompting. Given a prompt $\mathcal{P}$, the LLM input can be formed through pairs of knowledge $\mathcal{K}$ and queries $\mathcal{Q}$ resulting in $\mathcal{P} = \{\mathcal{K}, \mathcal{Q}\}$. This is used in RAG applications to append full documents or knowledge triples \cite{lewis2020retrieval, sun2023think}.

Such an approach is problematic as the LLM output depends on hand-crafting the prompt template through the overall phrasing of the query, quality of the relevant evidence, fixed context window lengths and lack of control over efficient usage of the prompt text by the model. These problems are also outlined by \cite{whatSota} and we support the call for more robust evaluation methodologies atleast through multi-prompt evaluations. To this end, reliance on prompting can also be observed in other previous works \cite{queryInYourTongue, betterAskEnglish, politicalExpertsKGsLLMs}. 

With respect to Table \ref{tab:datasets}, the only dataset that provides such multi-prompt evaluation is DefAn \cite{rahman2024defan}, which is a QA dataset where each data point is accompanied by ten different rephrasings of a question. Prompt-based knowledge injection is also limited by the context window size and does not deal with cases where the model's internal knowledge may conflict with the provided evidence. Therefore, context-aware decoding \cite{shi2023trusting} proposes a strategy for prioritizing in-prompt knowledge through a learnable parameter. It is worth noting that context-aware decoding requires two inference passes to generate a final output, therefore increasing the computational cost twofold.

%A recent contribution from the semantic web community 
Recently, \cite{lageweg2024generative} proposed a method for generating S-expressions based on extracted entities from a knowledge graph given a user query. The method is fully grounded on the data available in the knowledge graph, therefore while improving factuality, there are open questions towards generalizability and supporting cases, where KG data is incomplete or missing. 

Additionally, knowledge integration via adapter networks has been explored. \cite{tian-etal-2024-kg}, for example, proposes a method for dynamically injecting knowledge graph information in the latent space. This method is supported as it allows to encode rich metadata of KG triples which otherwise cannot be done at scale with prompting, and it enables rapid knowledge updates.

We hypothesize that a reliable LLM system development could contain a mixture of these mitigation strategies although it is unclear to what extent different methods complement one another.

%\subsection{Post-Generation}
\bigskip
\textbf{Post-Generation. }
Another line of work \cite{guan2024mitigating} proposes retrofitting LLM output factuality by consulting an external KG once an answer is generated by an LLM. The methodology follows a 5-stage pipeline, where an output is generated, claims are extracted, cross-checked against an external KG, and afterwards the original output is patched up as needed according to a claim verification module. Each of the five stages in the pipeline relies on an LLM performing the designated task.

%\subsection{Multilinguality}
\bigskip
\textbf{Multilinguality. }
Recent studies suggest that hallucinations are more prone in lower-resource languages \citep{chataigner2024multilingual}, and language models can have inconsistent knowledge representations \citep{qi2023cross} and disparities \cite{betterAskEnglish} across languages. Additionally, \cite{russaMultiling} outlines that multilingual KGs can be particularly useful for low-resource languages, where training data is limited with applications towards question answering, fact extraction, and others. Therefore, we identify multilingual knowledge integration as a necessary research direction that can be supported by reliable multilingual KGs, we refer the reader to a previous survey for details regarding multilingual KGs themselves \cite{russaMultiling}.

A previous method \cite{hou2022adapters} was proposed to improve a multilingual language model by statically encoding knowledge of a multilingual KG through a set of adapters. This results in helping the model align entities and knowledge in a multilingual space, thus making the internal knowledge representations more language-agnostic. The results of this work suggest mutual benefits for KGs and LLMs with applications on KG completion and KG entity alignment, as well as language understanding tasks, such as named entity recognition and question answering.

% todo: outline typological diversity maybe? Esthers paper

%\subsection{Limitations}
\bigskip
\textbf{Limitations. }
%\label{sect:limitations}
Reliance on knowledge pre-training means that the knowledge is encoded statically. While the methods suggest factual and task-specific improvements, this approach does not solve the fundamental problem of rapid knowledge updates required by use cases where knowledge develops continuously. Additionally, common reliance on LLMs and prompting during knowledge integration, inference, and post-generation gives more room for error, especially as the number of submodules involved in the processing pipelines grows. Prompting is also limited due to fixed context-window lengths, fragility of the handcrafted prompt templates and lack of control over the model's usage of the prompt.  This creates a trade-off for system designers for balancing between expensive inference passes for potential factuality gains. 

Similarly, as for hallucination detection, we call for in-depth reporting on the error rate of each submodule, as well as researching methodologies that move away from solving subtasks based on textual prompting. Additionally, we note that there are different types of fundamental approaches to using KGs for hallucination mitigation, such as incorporating them as part of pretraining, inference, or using KGs to retrofit LLM outputs. Therefore, we also call for research that explores the effects of stacking these approaches together and investigating the extent to which the different methodologies complement one another when used together. We also outline the importance of multilinguality.

\section{On Hallucination Evaluation}
\label{sect:hallcEval}
% Changing section title: scratch "evaluation"
% Consistent use of hallucinations over factualty
Evaluation of LLM factuality can easily become very complex depending on how the task is framed. The MedHalt and TruthfulQA \cite{umapathi2023med, lin2021truthfulqa} datasets contain subtasks that model question answering evaluation as a multi-choice task, meaning that an LLM is required to choose an answer from a predefined list of answers. While experiments can still be designed around this by measuring correct and incorrect responses with and without added knowledge, an LLM can still choose a correct answer simply by chance thus not leading to quality insights on the quality of models' internal knowledge representations.

To this end, framing the problem as a generative task and evaluating the semantics can be seen as a more robust approach. Metrics such as BERTScore \cite{zhang2019bertscore} or BARTScore \cite{yuan2021bartscore} are employed, which are neural-based approaches. Other lines of work use auxiliary LLMs \cite{zheng2023judging, li2023halueval} to evaluate whether a given LLM's output is hallucinatory or satisfactory for correct answers, thus scoring either correctly or incorrectly. These approaches themselves are prone to errors as language models are prone to hallucinations, this is especially evident in low-resource language settings \cite{hallcDetMethMultiLing}. Therefore, we argue for the importance of human-based evaluation, in at least a subset of the results, to indicate reliability in a similar spirite as \cite{li2023halueval} as well as considering languages beyond English for hallucination evaluation.

A recent work \cite{factscore} proposes a factuality estimation method FactScore. The methodology is based on two core ideas: (i) break an LLM output into atomic facts, and (ii) compare an atomic fact with respect to external knowledge. Although the original contribution defines the methodology by comparing the atomic facts to Wikipedia articles through entailment, this can be expanded to KGs as the atomic fact extraction allows to compare overlaps with KG triples therefore quantifying factuality on a more fine-grained, interpretable manner. We call for research towards robust methods of claim extraction from LLM output, which would then allow to perform evaluations with respect to KGs as sources of ground truth.

% review https://arxiv.org/abs/2402.10496 multilingual hallc. eval.

% Recent studies suggests that hallucinations are more prone in lower-resource languages \cite{chataigner2024multilingual}, and language models can have inconsistent knowledge representation across languages \cite{qi2023cross}. Generally the best performing and highest resource language being English. For LLM systems to be accessible and inclusive, it is important to enable evaluation for other languages than English. With respect Table \ref{tab:datasets}, the only dataset that provides multilinguality is Shroom-2025 shared task. 

% Furthermore \cite{qi2023cross} outlines that language models can have inconsistent knowledge representations across languages. Assuming that factual knowledge is language agnostic, further research and evaluation in a multilingual space can benefit towards a better understanding of knowledge representations.

% Link to NLP multiliguality discussion? What works for english may not work other languages, this needs to be considered. English has loads of resources, other languages are being neglected. Making use of other languages to benefit English, or vice versa?

% ---

% Outline previous works and limitations:
% \begin{enumerate}
%     \item Hallucinations are more prone in low-resource languages
%     \item Current metrics and their limitations: neural metrics, LLM-as-judge, humans
%     \item Need for resources on 
% \end{enumerate}

\section{Conclusions and Future Work}
\label{sect:conclusions}
In this paper, we highlight and discussed the importance of using KGs as a potential solution for mitigating the hallucination phenomenon. By surveying current literature and analyzing the limitations, we identified useful research directions within resources, hallucination detection, and external knowledge integration. While previous methods suggest improvements, hallucination mitigation is still an ongoing research problem with no single solution that is general enough to solve the task at hand. We believe the semantic web and NLP communities together can solve the problem by combining expertise and research within 
%that the Semantic Web community can bring contributions towards the NLP field, especially through contributing expertise and research within 
effective and multilingual graph creation and completion, entity extraction from text, graph embedding extraction, multilingual entity linking, and exploring methods of synergizing KGs and LLMs.

We further identify the following future directions for research within hallucination mitigation in LLMs using knowledge graphs:
\begin{enumerate}
    \item Large-scale datasets that provide accurate KG triples as added context, including training, development, and test splits. This is necessary to further the research on parametric methods for knowledge integration. Additionally, if such a dataset includes KG triples for inputs, such a dataset can additionally be used for parametric entity extraction.
    \item Robust evaluation that includes multilinguality, multitasks, and multiprompts. This gives a better insight into how truly generalizable and robust a particular system can be. Such robust evaluation is generally not included in modern studies, where the evaluation is normally done using single prompts, single language, which normally is English, and in most cases a single task.
    \item Hallucination detection with a fine-grained overview of hallucinatory text spans. Hallucination detection is the first step for mitigating hallucinations, therefore robust knowledge within detection can greatly benefit mitigation.
    \item Knowledge integration methods that move away from textual prompt reliance, ideally in a parameter-efficient setting. This is supported by the fragility towards prompt formatting and comprehension, context window limitations.
    \item Studies on mixing and matching fundamentally different methods of hallucination mitigation methods. This can provide an insight into how methods complement one-another and can be particularly valuable for industry practitioners when designing systems.
    \item Multilinguality for hallucination detection, evaluation, and knowledge integration.
\end{enumerate}

\section*{Acknowledgements}
This work is supported by the Poul Due Jensens Fond (Grundfos Foundation).

% \bibliographystyle{elsarticle-num}
\bibliographystyle{elsarticle-num-names} 

\bibliography{bibliography.bib}



%% For example:

%% \bibitem[Aladro et al.(2015)]{Aladro15} Aladro, R., Martín, S., Riquelme, D., et al. 2015, \aas, 579, A101

%%\end{thebibliography}

\end{document}